\documentclass[11pt]{article}
\usepackage[preprint]{acl}
\usepackage[T1]{fontenc}
\usepackage[utf8]{inputenc}
\usepackage{times}
\usepackage{latexsym}
\usepackage{amsmath}
\usepackage{booktabs}
\usepackage{graphicx}
\usepackage{microtype}
\usepackage{url}
\usepackage{array}
\usepackage{tabularx}

\title{Novels generated by language models show compressed formal variation}

\author{Mehdy Sedaghat Payam \\
  University of Maryland \\
  \texttt{mspayam@umd.edu} \\
  \texttt{ms79payam@yahoo.com}
  \And
  Justin Quinn \\
  University of West Bohemia \\
  Charles University \\
  \texttt{jquinn@fpe.zcu.cz}}

\begin{document}
\maketitle

\begin{abstract}
While large language models can generate entire novels, there is little information about the level of formal variation in their output over many generations. Rather than asking whether individual passages can be identified as AI-generated, this study asks whether repeated AI generation can produce the same range of diversity which is found across human corpora. This paper contrasts six corpora based on generation source and target style: twenty novels generated using GPT-5.5 Thinking in a nineteenth-century British realist style, twenty novels generated using Qwen3-14B in a nineteenth-century British realist style, twenty novels generated using each of these models in a contemporary ``zero style,'' 205 nineteenth-century human-written British novels, and sixty-five contemporary human-written Zero-Style novels. At the document level, the research includes MATTR-500, Shannon entropy, average sentence length, readability, and punctuation rate measurements. The most robust and reliable result is compression of sentence structure. Repeated generations produce novels that vary far less from one another in sentence structure than human novels do. Compression is also present in the measures of readability, punctuation, and sentence length variability within novels. Lexical measures tend to be similarly compressed, with the exception of Qwen Zero-Style MATTR. Despite having distinct mean stylistic profiles, GPT and Qwen lack a stable pattern of cross-measure correlation. This article therefore distinguishes between variance overclosure, which represents a limited formal range between novels, and a more specific phenomenon of correlational overclosure. This means that an individual AI-generated novel may resemble human fiction stylistically, while a collection of AI-generated novels occupies a much narrower formal range.
\end{abstract}


\section{Introduction}

\subsection{From surface narration to corpus structure}

Large language models can now produce texts that, according to the German philosopher and literary scholar, Hannes Bajohr, are linguistically ``cohesive'' and stylistically ``unmarked.'' He uses ``surface narration'' to refer to this identifiable type of storytelling that emerges when cohesive sequences give an impression of cause-and-effect and time continuity. Although ``surface narration'' highlights an important element in the AI narratives, it focuses on the connection between cohesion and coherence in a story, not the forensic detection of machine authorship, which is the domain of stylometry (Bajohr 2024).

Stylometry research in this area has found that human-written texts and short texts generated by AI retain differences in terms of their lexicon, syntax, grammar, and punctuation (Kobak et al. 2025; O'Sullivan 2025; Pawłowski and Walkowiak 2026; Przystalski et al. 2026). This paper raises another question that these studies do not address: \textbf{Do formal differences persist at the scale of complete novels, where LLMs must repeatedly sustain patterns across tens of thousands of words?}

A novel is more than just an extended short text. It requires the model to make stylistic decisions repeatedly about lexical renewal, rhythm of sentences, readability, punctuation, and their interplay\footnote{Sometimes writers select a particular set of these features based on deliberate, predetermined decisions in order to construct their personal style (Hemingway, for example), and critics such as William Gass (1985, p. 150) classify writers according to these predetermined decisions.}. A novel produced by the model can be fluent and stylistically varied; yet a set of generated novels may cover far less formal variation than a set of human-written novels. What counts here is not the individual sentence or passage, but how varied the novels are from each other and whether formal properties remain independent.

This article does not aim to determine whether the author of a short passage is AI or a human. Instead, the goal is to see whether the corpus-wide diversity of AI-written novels mirrors that of two selected human comparators. The study will also assess whether the patterns noted are restricted to a particular generation system or can be seen across different model setups. In order to do so, this study examines novels written using GPT-5.5 Thinking via the ChatGPT interface and novels written via Qwen3-14B through a separately documented Google Colab workflow.

\subsection{Contribution beyond stylometric detection}

The closest similar study by O'Sullivan (2025) compares human-generated and AI-generated creative writing and shows that the style of short generated text can be distinguished from human writing. There are five differences between the current study and O'Sullivan. First, the current study focuses on complete novels rather than short creative texts. Second, each novel is taken as an individual observation at the document level rather than being classified as to its type. Third, the current study uses the dispersion across novels and the coordination of features as its dependent variables. Fourth, the uncertainty is estimated by matching the size of resamples from both the generated and human groups in each generated-human comparison. Fifth, the current study explores whether the major findings hold for two different model settings rather than focusing on the outputs of one particular model interface workflow.

\subsection{Homogenization across textual scales}

Recent research has identified convergence in AI-generated cultural production at several textual and social scales. Anderson et al. (2024) study homogenization in the ideation process when people are aided by language models. Agarwal et al. (2025) demonstrate that AI writing recommendations contribute to stylistic convergence both within and between cultural communities and change culturally determined writing styles towards Western ones. As for the narrative level, Rettberg and Wigers (2025) observe that stories created for various national contexts always repeat similar plots of homecoming, tradition, community organizing, and reconciliation. Similarly, Wilkens (2026) observes that homogeneity is based on generation design; simple prompts yielded homogenous chapters, while complex prompts with personalized synthetic biographies generated diversity in the embedding space. These studies are concerned with related but different phenomena. Here, we focus on another aspect of text organization: the formal distribution that arises when multiple novels are taken into consideration. Do generated novels form compressed distributions in terms of document-level stylistic features? Do features that are relatively independent in human fiction become more coordinated in generated texts?

The presence of both GPT-5.5 Thinking and Qwen3-14B allows us to establish whether these tendencies occur solely within one configuration of the model--interface or are common for two different generation processes. Consistency between the two models would mean that formal compression is not exclusive to ChatGPT. However, the consistency itself would not prove the universality or inherent nature of overclosure in language models. The findings are still limited to the specific models, interfaces, prompts, topics, and generation processes considered.

Formal range, then, is an analytical perspective. It is broader than the local coherence of particular passages but does not cover the entire range of plot, character development, focalization, cultural depiction, or narrative significance.

\subsection{Formal range and overclosure}

Human literary corpora differ not only in the average values of stylistic features, but also in the degree of dispersion and the relationships among these features (Eder 2017; Eder et al. 2016; Stamatatos 2009). For example, two authors might display similar levels of lexical diversity through different sentence structures, or two novels with nearly identical readability might use punctuation and vocabulary in different ways. This relative independence gives literary corpora a diverse formal range. In this article, when we say ``formal range,'' we specifically mean the dispersion and coordination among the stylistic features we have measured; we do not mean plot, characterization, point of view, or overall narrative structure.

Overclosure describes a corpus-level condition in which generated texts remain locally fluent and coherent but depict an unusually restricted tendency across the measured formal dimensions. ``Variance overclosure'' means that novels differ very little from one another with respect to features such as lexical diversity, entropy, sentence length, or readability. ``Correlational overclosure'' is an exploratory hypothesis concerning features that are not mechanically related but become more strongly coordinated in generated texts than in a human comparator.

These two types of overclosure differ from one another: a corpus can be restricted without its variables becoming interdependent, or conversely, its variables can become interdependent without all of them being compressed to the same degree. Overclosure does not mean that a text is simple, low-quality, or lexically weak. A corpus might have a high average lexical diversity while still showing very little difference between its novels. Nor does this term describe a single text on its own; rather, it describes a pattern seen across an entire set of texts produced under similar conditions.

\subsection{Research questions}

RQ1. Do the four generated conditions exhibit lower inter-novel dispersion than their corresponding human comparators?

RQ2. Does any observed compression persist after document-count, authorship, period, and length controls?

RQ3. Do GPT and Qwen occupy similar or different mean stylistic locations within each stylistic target?

RQ4. Do cross-feature correlations display a replicable pattern across models and stylistic conditions?

The main comparison is at the corpus level, where each generated condition is compared to the corresponding human corpus. The design does not try to estimate any general effect of AI-generated works, of prompting, or of model architecture. Instead, it tries to estimate the formal range created by a repeated model--workflow combination in comparison with the formal range of a chosen human multi-author corpus. It does not try to estimate whether AI-generated novels have lower variability than the works of a single human author.

The analysis distinguishes between three different levels of quantitative evidence. The first one is whether the generated novels demonstrate less inter-novel variance in terms of the formal features that have been measured. The second one is related to comparing the mean of stylistic positions of GPT and Qwen. Finally, the cross-feature correlations are treated as an exploratory analysis of whether there is any pattern of coordination between features for all models and stylistic targets. The close reading part is the qualitative equivalent of this analysis, which examines whether the corpus compression can be detected in individual novels.

\section{Corpora and methods}

\subsection{Six-corpus comparative design}

This research makes use of six corpora that are classified into two main conceptual dimensions: generation source and stylistic target. In terms of generation source, there is a differentiation between human-written novels and the ones written by GPT-5.5 Thinking and Qwen3-14B. In terms of stylistic target, there is a distinction between fiction written in the style of the nineteenth century (Victorian novels) and novels written in a contemporary style that is intended to be accessible and has low stylistic markedness. This results in two human comparators and four generated conditions.

\begin{table*}[t]
\centering
\small
\caption{Corpus composition and document-length distribution.}
\label{tab:1}
\renewcommand{\arraystretch}{1.08}
\resizebox{\textwidth}{!}{%
\begin{tabular}{@{}lllllll@{}}
\toprule
\textbf{Corpus} & \textbf{N} & \textbf{Authors / generator} & \textbf{Mean words} & \textbf{Median} & \textbf{Word-count SD} & \textbf{Role} \\
Human-19th Century & 205 & 121 human authors & 142,618 & 142,739 & 81,418 & Historical comparator, 1800-1900 \\
GPT Victorian & 20 & GPT-5.5 Thinking / ChatGPT & 160,576 & 176,954 & 40,023 & Victorian-style GPT condition \\
Qwen Victorian & 20 & Qwen3-14B / Colab L4 & 51,885 & 52,432 & 1,449 & Victorian Qwen condition \\
Human Zero-Style & 65 & 29 human authors & 77,751 & 70,343 & 36,133 & Accessible contemporary comparator \\
GPT Zero-Style & 20 & GPT-5.5 Thinking / ChatGPT & 107,960 & 103,481 & 13,222 & Zero-Style GPT condition \\
Qwen Zero-Style & 20 & Qwen3-14B / Colab L4 & 51,261 & 51,296 & 1,557 & Zero-Style Qwen condition \\
\bottomrule
\end{tabular}%
}
\end{table*}

\subsection{Nineteenth-Century British Human corpus}

The historical comparison corpus is drawn from the ``Victorian Novels'' collection, made available through a project called Ciphers of the Times (2022). After setting aside texts from the twentieth century, 205 works remained, published between 1800 and 1900. Because this time span begins before Queen Victoria's accession to the throne, we have named this corpus ``Nineteenth-Century British Human'' rather than ``Victorian Human.'' This collection includes works by 121 authors and spans different decades, various subgenres, different narrative styles, different text lengths, and different levels of recognition and literary standing.

This is a historically broad and specific sample and should not be considered a universal standard for the value of human literature. The breadth of this corpus allows us to measure the degree of compression in the texts; meanwhile, it should be noted that this corpus is shaped by factors such as how books were digitized, the availability of works in the public domain, and which works have come to be recognized as important over time (Ciphers of the Times 2022; Jockers 2013; Moretti 2013; Underwood 2019).

\subsection{Zero-Style Human Corpus}

Contemporary human corpus consists of sixty-five novels selected by Justin Quinn according to his definition of zero style, which refers to prose aiming at global readability and linguistic accessibility, as well as low stylistic markedness (Quinn 2025). The full list of titles is given in Supplementary Table S2.

Although this corpus contains writings of several writers, not all writers have an equal number of novels included in this corpus; for instance, Ha Jin's corpus consists of eight novels, and the corpus of Xiaolu Guo comprises six novels, while for some writers only one novel was selected. The size-matched human subsampling used below reduces the effect of one writer on the whole corpus.

\subsection{GPT Corpora}

The GPT corpora were generated with GPT-5.5 Thinking through the ChatGPT interface. This process started in early May 2026 and finished in mid-June 2026, with roughly one novel produced per day. The generation of these novels was not a single-step process but was carried out interactively and continuously: a chapter-by-chapter outline and the first chapter of the novel were requested first. Then the user obtained the rest of the novel across several successive turns using continuation commands. The information about the user interface, such as the model' s creativity setting (temperature), the top-p value, the random seed used for sampling, the exact version of the backend system, the system prompt, or how context and memory were managed was not revealed. For this reason, this research is merely a combination of the model, the user interface, and the prompting method, not the language model alone. None of the conclusions of this article depend on knowing these hidden parameters, and we do not attribute the patterns we observed to any specific internal mechanism of the model.

The GPT Victorian prompts called for the following characteristics: standard nineteenth-century British English, omniscient narration that occasionally uses the technique of ``free indirect discourse,'' detailed and realistic description of social interactions between characters, and avoidance of directly imitating the style of specific named authors or using anything from after 1900. These prompts specifically requested that sentences vary in length and rhythm, meaning both long, multi-part sentences and shorter sentences should appear side by side. Additionally, the social world of the stories had to be built around themes such as inheritance, marriage, work, education, social respectability, religious or political disputes, and generational conflict. In the prompt archive, there is a general template for long-form novels aiming for 35 to 65 chapters and about 120,000 words. We report here the actual length of each novel rather than assuming that all novels reached exactly one fixed target. We have also taken the effect of novel length into account and controlled for it in the statistical analysis.

The ``Zero-Style'' prompts called for the following characteristics: contemporary English suitable for readers around the world, text that non-native English speakers could read comfortably, a consistent and uniform narrative voice throughout the story, no imitation of any specific author's style, and no use of experimental forms, stream of consciousness, or unusual typographic play with the writing and page layout. The narrator could be first-person or third-person. Each prompt asked for a novel of 15 to 30 chapters and about 60,000 to 80,000 words. The twenty different stories in this collection take place in the 1990s, 2000s, and 2010s, and are set in various countries in North America, Europe, South Africa, Australia, and New Zealand. However, these stories share one common point: they are all about migration, life in diaspora, institutions situated between multiple cultures, or minority communities. In Supplementary File 1, both the shared patterns across the prompts and the specific sections of each individual prompt are presented.

\subsection{Qwen3-14B Corpora}

Qwen3-14B is a strong mid-size open-weight language model with 14.8 billion parameters created by Alibaba's Qwen Team. To examine whether the pattern was confined to the initial GPT configuration, a second corpus was generated using an independently configured Qwen workflow. Because model, prompting, segmentation, context management, and decoding conditions differed simultaneously, this comparison is treated as cross-workflow robustness rather than a controlled model comparison.

The Qwen corpora were created in June 2026 using the Qwen/Qwen3-14B checkpoint (Yang et al. 2025) in Google Colab Pro L4 runtime. Twenty pairs of story seeds were designed; each seed was generated once under a Victorian lock and once under a zero lock. The order of generation was balanced for each pair. Each novel aimed for twenty chapters and 60,000 words. Each story seed was used once under each stylistic condition, making comparisons between the two Qwen conditions paired rather than fully independent.

Novel generation was divided into segments. Generation calls were for a target of about 700 words, with no more than 1,200 tokens, accompanied by 1,400 words of recent prose along with a concise continuity state. The input size limit was 20,000 tokens. Prose generation used a temperature 0.70, top-p 0.80, top-k 20, and repetition penalty 1.20. Planning calls used a lower temperature and repetition penalty. There was an online duplicate detection filter that would reject segments containing exact duplicates within the same novel.

The Qwen Victorian condition employed a British realist perspective of the period 1830--1890, third-person point of view, historically limited institutions, and social conflicts similar to those in the GPT Victorian target. The Qwen Zero-Style condition employed realism, third-person point of view, and clear international English.

The use of the fixed target segments, chapter structure, continuity summaries, context window, and decoding options makes this workflow reproducible, but it might also regularize its outputs; thus, the output can be attributed to the entire model-workflow configuration and not just Qwen's architecture. Although treated as document-level observations, these novels are repeated outputs from shared configurations rather than unrestricted independent draws from the population of all possible model outputs.

\subsection{Text processing and measures}

All the novels were examined and analyzed using the same computer program written in the Python language. To calculate the lexical features of the text, all English words (including words with apostrophes) were converted to lowercase letters. To measure word diversity, a method called MATTR was used, which examines consecutive words in small windows (500 words for this research) at a time and then calculates their average (Covington and McFall 2010).

To measure the distributional diversity of word use across each complete text, a mathematical formula called ``Shannon entropy'' was used, which is calculated based on the repetition of words in each text (Shannon 1948; Tanaka-Ishii and Aihara 2015). The average sentence length was also calculated based on the number of words per sentence, after the sentences were split in a uniform way.

For the readability score, eight standard formulas were applied. In the main text, only the Flesch Reading Ease formula is presented since this scale can be easily interpreted and understood due to its reliance on sentence length and number of syllables per word (Flesch 1948). The other seven formulas are reported in the supplementary data. Since readability formulas traditionally only capture some surface characteristics but do not fully capture coherence and discourse processing, as well as readers' knowledge and literary quality, Flesch should be understood as a relative measure of readability (Crossley et al. 2008; Pitler and Nenkova 2008). Because sentence length is an explicit component of Flesch Reading Ease, readability compression is treated as supporting rather than independent evidence for sentence-structural regularization.

Finally, the number of punctuation marks (such as commas and periods) was also counted, and in order to be able to compare texts of different lengths with one another, this count was calculated and standardized per 1,000 characters of text. Treating punctuation as an element of style dates back to the Prague School; however, this paper follows a more recent approach that examines patterns of punctuation across literary texts (Vachek 1989; Darmon et al. 2021).

Each of these tools measures something different. MATTR, which estimates local lexical diversity, is designed not to depend too heavily on the overall length of the text (Covington and McFall 2010). Entropy shows how evenly words are distributed throughout the whole text, but it may still be affected by text length or the number of different words (Shannon 1948; Tanaka-Ishii and Aihara 2015). Average sentence length measures the rhythm and pacing of sentences, but on its own it cannot represent the full complexity of sentence structure (Biber and Conrad 2009). Readability formulas provide only rough estimates of how easily a text can be read; they do not measure literary quality (Flesch 1948; Gunning 1952). Punctuation provides supplementary information about clause segmentation and the rhetorical organization of written prose (Nunberg 1990; Biber and Conrad 2009).

\subsection{Operationalizing variance and correlational overclosure}

To check ``variance overclosure,'' we compare the standard deviation of the AI-generated texts with the standard deviation of the human texts. If this ratio is less than 1, it means the difference among the generated novels is smaller than the difference among the human novels. This ratio is just a simple description; to find out whether this difference is meaningful, we use the Brown--Forsythe test,\footnote{The Brown--Forsythe test checks whether the corpora differ in how much their values vary and is less affected by unusual values or uneven distributions} as well as a resampling method\footnote{The author-balanced bootstrap repeatedly resamples generated novels and human authors, drawing one novel for each sampled human author, so that each iteration compares equally sized sets without allowing prolific authors to dominate the human sample.}, and length-adjusted residual analyses\footnote{The residual analysis adjusts the measurements for differences in novel length and then checks whether the AI novels still show less variation than the human novels.}.

Correlational overclosure is assessed only for feature pairs that are not mechanically coupled by their definitions. For example, the correlation between sentence length and readability formulas like Flesch cannot be taken as a sign of anything special, since sentence length is itself part of how those formulas are built. Similarly, a high correlation between MATTR and entropy is also expected, since both are derived from word frequency; this correlation is not evidence of some AI-specific uniformity. So the two important relationships worth examining are, first, the relationship between MATTR and average sentence length, and, second, the relationship between entropy and average sentence length.

\subsection{Statistical procedures}

First, the Brown--Forsythe tests examine the variance of documents for each generated condition in comparison with the corresponding human one (Brown and Forsythe 1974). The main family of inferential tests includes sixteen tests: four measures for four pairs of generated and human conditions. The Benjamini-Hochberg correction is applied to the false discovery rate (FDR)\footnote{The false discovery rate (FDR) correction reduces the chance of treating random results as significant when several statistical tests are conducted at the same time.} for these tests (Benjamini and Hochberg 1995).

Second, a two-sided author-balanced bootstrap (Efron and Tibshirani 1993) assesses the uncertainty in the dispersion ratio. In each of 10,000 repetitions, twenty novels generated by the model are sampled with replacement. For the human corpus, twenty authors are sampled with replacement, after which one novel is randomly chosen for each sampled author. For every repetition, the ratio of the standard deviation of the generated corpus to the standard deviation of the human corpus is calculated. We present the median ratio, its 95\% percentile interval, and the percentage of repetitions in which generated dispersion is higher than human dispersion. This procedure equalizes the number of documents on the two sides while preventing prolific human authors from receiving greater weight simply because more of their novels are included in the corpus.

Third, we take into account differences in the length of the documents. For every measure, the model estimates the expected value of the measure based on the size of the novel being analyzed. The difference between the observed value and the estimated one is considered the residual. Then, we check how these residuals vary in both the human and artificial corpora.

Fourth, sensitivity analyses included leave-one-author-out tests, joint removal of the two most represented Zero-Style authors, author-equalized resampling using one novel per author, restriction of the historical corpus to 1837--1900, and within-author calibration for human authors represented by at least three novels.

Finally, cross feature co-analysis was carried out using author-balanced comparisons of Pearson correlations after Fisher's z-transformation, and Spearman correlations were used as a rank-based check. Document-level dispersion was the primary analysis; author-balanced resampling and length adjustment were robustness analyses; the human-comparator checks were sensitivity analyses; and correlations and close reading were exploratory. All Brown--Forsythe tests were two-sided, with an unadjusted alpha of .05. The sixteen primary comparisons formed one correction family, with Benjamini--Hochberg adjustment applied across that family.

\section{Results}

\subsection{Primary distributions and variance compression}

\begin{table*}[t]
\centering
\small
\caption{Document-level means, standard deviations, raw dispersion ratios, and FDR-adjusted Brown-Forsythe tests.}
\label{tab:2}
\renewcommand{\arraystretch}{1.08}
\resizebox{\textwidth}{!}{%
\begin{tabular}{@{}llllll@{}}
\toprule
\textbf{Comparison} & \textbf{Measure} & \textbf{Generated mean (SD)} & \textbf{Human mean (SD)} & \textbf{SD ratio} & \textbf{FDR q} \\
GPT Victorian / 19C Human & MATTR-500 & 0.577 (0.012) & 0.513 (0.022) & 0.536 & .017 \\
GPT Victorian / 19C Human & Shannon entropy & 9.762 (0.131) & 9.356 (0.251) & 0.523 & .008 \\
GPT Victorian / 19C Human & Mean sentence length & 14.440 (0.639) & 23.936 (10.808) & 0.059 & .008 \\
GPT Victorian / 19C Human & Flesch Reading Ease & 66.857 (2.702) & 66.965 (9.916) & 0.273 & .004 \\
Qwen Victorian / 19C Human & MATTR-500 & 0.517 (0.007) & 0.513 (0.022) & 0.325 & $<$.001 \\
Qwen Victorian / 19C Human & Shannon entropy & 8.804 (0.069) & 9.356 (0.251) & 0.276 & $<$.001 \\
Qwen Victorian / 19C Human & Mean sentence length & 16.513 (0.638) & 23.936 (10.808) & 0.059 & .008 \\
Qwen Victorian / 19C Human & Flesch Reading Ease & 77.556 (1.145) & 66.965 (9.916) & 0.115 & $<$.001 \\
GPT Zero-Style / Quinn Human & MATTR-500 & 0.543 (0.011) & 0.521 (0.028) & 0.387 & .004 \\
GPT Zero-Style / Quinn Human & Shannon entropy & 9.401 (0.086) & 9.390 (0.263) & 0.326 & $<$.001 \\
GPT Zero-Style / Quinn Human & Mean sentence length & 10.597 (0.331) & 16.441 (4.916) & 0.067 & .003 \\
GPT Zero-Style / Quinn Human & Flesch Reading Ease & 75.932 (3.370) & 75.314 (6.977) & 0.483 & .045 \\
Qwen Zero-Style / Quinn Human & MATTR-500 & 0.509 (0.016) & 0.521 (0.028) & 0.576 & .073 \\
Qwen Zero-Style / Quinn Human & Shannon entropy & 8.762 (0.146) & 9.390 (0.263) & 0.554 & .032 \\
Qwen Zero-Style / Quinn Human & Mean sentence length & 13.453 (0.569) & 16.441 (4.916) & 0.116 & .004 \\
Qwen Zero-Style / Quinn Human & Flesch Reading Ease & 80.301 (1.885) & 75.314 (6.977) & 0.270 & .005 \\
\bottomrule
\end{tabular}%
}
\end{table*}

SD ratio = generated-corpus SD / matched human-corpus SD

The principal pattern in Table 2 is reduced inter-novel dispersion rather than convergence on a shared mean stylistic profile. Of the sixteen raw ratios of AI-generated to human SD, fifteen are under 0.58, and fifteen of the Brown-Forsythe comparisons are still under q = .05 even after adjustment. The one exception is Qwen Zero-Style MATTR, where the ratio is 0.576 and the adjusted q is .073.

Readability scores show why we need to distinguish between the mean and the variance. The average Flesch score of the Zero-Style AI corpus (75.93) and the Zero-Style human corpus (75.33) is almost equal, but the range of Flesch score of the AI corpus is much smaller than that of the human one. This is also true in terms of entropy.

Intra-novel sentence-length dispersion reinforces the result. The between-novel SD ratios for the standard deviation of sentence length are 0.051 for GPT Victorian, 0.034 for Qwen Victorian, 0.052 for GPT Zero-Style, and 0.089 for Qwen Zero-Style. Generated novels therefore converge not only in their mean sentence length but also in the amount of sentence-length variation they sustain internally.

\subsection{Author-balanced bootstrap and robustness checks}

The following analyses test whether the observed variance compression can be explained by unequal corpus sizes, document length, repeated human authorship, or the historical breadth of the nineteenth-century comparator.

\begin{table*}[t]
\centering
\small
\caption{Two-sided author-balanced bootstrap of inter-novel standard-deviation ratios.}
\label{tab:3}
\renewcommand{\arraystretch}{1.08}
\resizebox{\textwidth}{!}{%
\begin{tabular}{@{}lllll@{}}
\toprule
\textbf{Comparison} & \textbf{Measure} & \textbf{Median SD ratio} & \textbf{95\% percentile interval} & \textbf{Generated SD \textgreater{} human SD} \\
GPT Victorian / 19C Human & MATTR-500 & 0.526 & 0.356--0.809 & 0.28\% \\
GPT Victorian / 19C Human & Shannon entropy & 0.498 & 0.315--0.794 & 0.27\% \\
GPT Victorian / 19C Human & Mean sentence length & 0.073 & 0.022--0.128 & 0.00\% \\
GPT Victorian / 19C Human & Flesch Reading Ease & 0.252 & 0.157--0.418 & 0.00\% \\
Qwen Victorian / 19C Human & MATTR-500 & 0.316 & 0.195--0.505 & 0.00\% \\
Qwen Victorian / 19C Human & Shannon entropy & 0.264 & 0.169--0.418 & 0.00\% \\
Qwen Victorian / 19C Human & Mean sentence length & 0.074 & 0.022--0.127 & 0.00\% \\
Qwen Victorian / 19C Human & Flesch Reading Ease & 0.106 & 0.064--0.177 & 0.00\% \\
GPT Zero-Style / Quinn Human & MATTR-500 & 0.389 & 0.234--0.661 & 0.00\% \\
GPT Zero-Style / Quinn Human & Shannon entropy & 0.322 & 0.211--0.506 & 0.00\% \\
GPT Zero-Style / Quinn Human & Mean sentence length & 0.062 & 0.034--0.144 & 0.00\% \\
GPT Zero-Style / Quinn Human & Flesch Reading Ease & 0.439 & 0.250--0.858 & 0.86\% \\
Qwen Zero-Style / Quinn Human & MATTR-500 & 0.583 & 0.396--0.953 & 1.67\% \\
Qwen Zero-Style / Quinn Human & Shannon entropy & 0.548 & 0.383--0.830 & 0.34\% \\
Qwen Zero-Style / Quinn Human & Mean sentence length & 0.107 & 0.051--0.257 & 0.00\% \\
Qwen Zero-Style / Quinn Human & Flesch Reading Ease & 0.244 & 0.126--0.489 & 0.01\% \\
\bottomrule
\end{tabular}%
}
\end{table*}

\emph{Note: 10,000 two-sided resamples. Generated documents and human authors were sampled with replacement; one novel was selected for each sampled human-author occurrence.}

\begin{figure*}[t]
\centering
\includegraphics[width=0.72\textwidth]{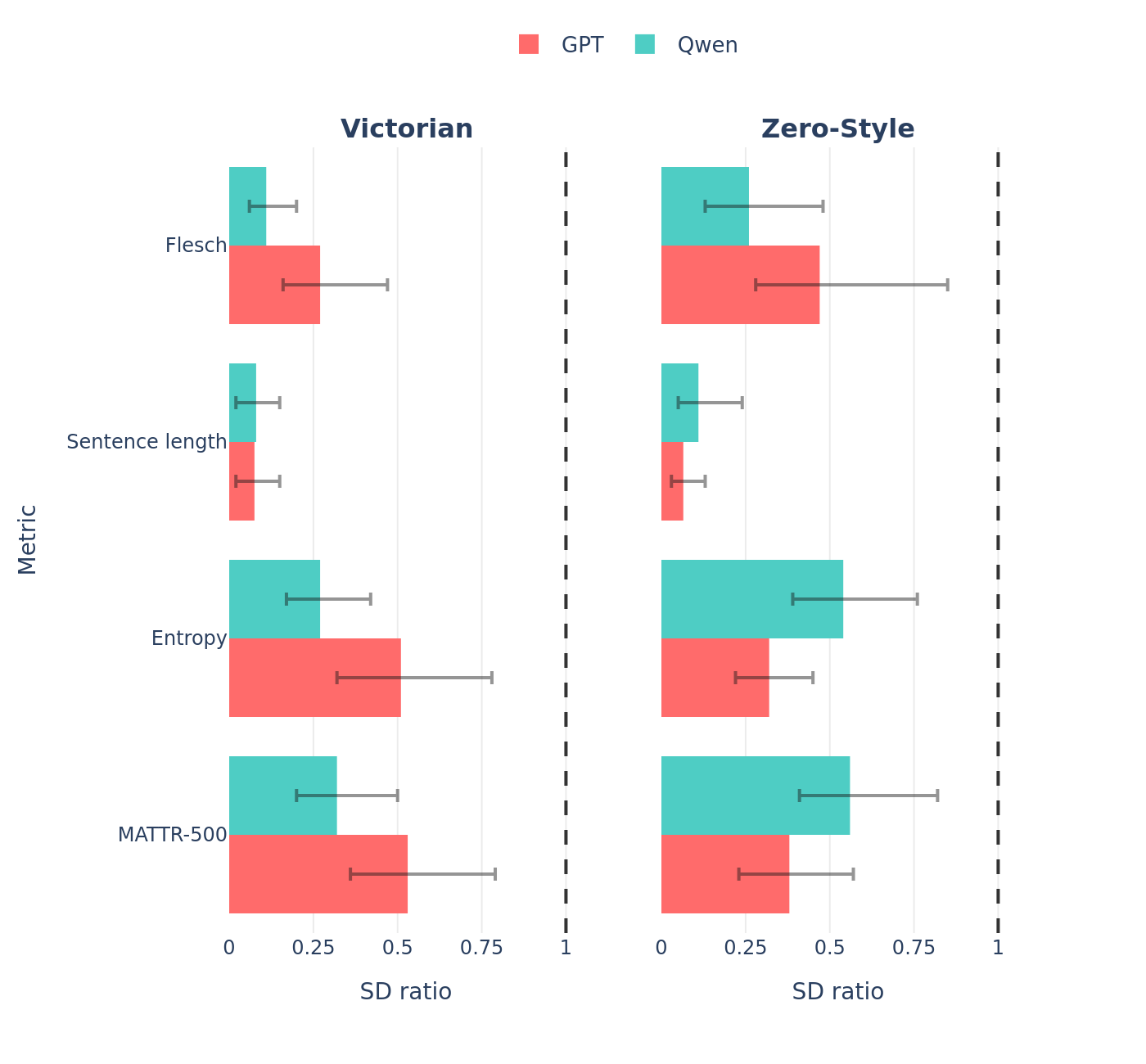}
\caption{Median two-sided author-balanced bootstrap ratios and 95\% percentile intervals. The dashed reference line at 1 marks equal inter-novel dispersion. All intervals remain below 1.}
\label{fig:bootstrap}
\end{figure*}

The Qwen workflow shows the same direction of variance compression under a substantially different generation process, while retaining distinct lexical means, readability levels, and punctuation patterns.

\subsection{}\label{section}

\begin{table*}[t]
\centering
\small
\caption{Residual dispersion after adjustment for log word count and corpus membership.}
\label{tab:4}
\renewcommand{\arraystretch}{1.08}
\resizebox{\textwidth}{!}{%
\begin{tabular}{@{}llllll@{}}
\toprule
\textbf{Comparison} & \textbf{Measure} & \textbf{Generated residual SD} & \textbf{Human residual SD} & \textbf{Residual ratio} & \textbf{FDR q} \\
GPT Victorian / 19C Human & MATTR-500 & 0.0117 & 0.0222 & 0.528 & .017 \\
GPT Victorian / 19C Human & Shannon entropy & 0.1056 & 0.2135 & 0.495 & .011 \\
GPT Victorian / 19C Human & Mean sentence length & 0.7795 & 10.7740 & 0.072 & .009 \\
GPT Victorian / 19C Human & Flesch Reading Ease & 3.0632 & 9.7850 & 0.313 & .005 \\
Qwen Victorian / 19C Human & MATTR-500 & 0.0072 & 0.0222 & 0.325 & $<$.001 \\
Qwen Victorian / 19C Human & Shannon entropy & 0.0681 & 0.2135 & 0.319 & $<$.001 \\
Qwen Victorian / 19C Human & Mean sentence length & 0.6399 & 10.7739 & 0.059 & .009 \\
Qwen Victorian / 19C Human & Flesch Reading Ease & 1.1494 & 9.7847 & 0.117 & $<$.001 \\
GPT Zero-Style / Quinn Human & MATTR-500 & 0.0112 & 0.0273 & 0.412 & .004 \\
GPT Zero-Style / Quinn Human & Shannon entropy & 0.0902 & 0.2382 & 0.378 & $<$.001 \\
GPT Zero-Style / Quinn Human & Mean sentence length & 0.2990 & 4.8612 & 0.062 & .002 \\
GPT Zero-Style / Quinn Human & Flesch Reading Ease & 3.3497 & 6.9270 & 0.484 & .032 \\
Qwen Zero-Style / Quinn Human & MATTR-500 & 0.0159 & 0.0273 & 0.584 & .077 \\
Qwen Zero-Style / Quinn Human & Shannon entropy & 0.1438 & 0.2382 & 0.604 & .036 \\
Qwen Zero-Style / Quinn Human & Mean sentence length & 0.5641 & 4.8612 & 0.116 & .003 \\
Qwen Zero-Style / Quinn Human & Flesch Reading Ease & 1.8930 & 6.9270 & 0.273 & .003 \\
\bottomrule
\end{tabular}%
}
\end{table*}

Adjustment for document length does not change the main trend. Residual ratios of sentence lengths are 0.072 for GPT Victorian, 0.059 for Qwen Victorian, 0.062 for GPT Zero-Style, and 0.116 for Qwen Zero-Style. All adjusted readability ratios remain under 0.49. Qwen Zero-Style MATTR remains the weakest lexical case (ratio 0.584, q = .077), while its entropy residuals remain compressed (ratio 0.604, q = .036).

The Qwen corpora exhibit extremely tight word counts, averaging 51,885 words for Qwen Victorian and 51,261 words for Qwen Zero-Style. As the residual analysis involves computing the expected value of the metric based on the length of the document, the persistence of the sentence length and readability metrics demonstrates that the compression of the Qwen corpora is not simply an incidental result of their shorter documents.

Leave-one-author-out sensitivity likewise supported the raw dispersion result. Removing each of the 121 historical and 29 Zero-Style authors in turn and recalculating the generated-to-human SD ratio produced sentence-length ranges of 0.058--0.083 for both Victorian comparisons, 0.064--0.080 for GPT Zero-Style, and 0.110--0.138 for Qwen Zero-Style. Every calculated ratio across the four measures remained below 1.

Restricting the historical comparator to 152 novels by 87 authors published from 1837 to 1900 produced the same direction under the implemented author-balanced bootstrap. The median sentence-length SD ratios were 0.102 for GPT Victorian (95\% percentile interval 0.063--0.169) and 0.101 for Qwen Victorian (0.064--0.165); all eight intervals across MATTR-500, entropy, sentence length, and Flesch remained below 1.

\subsection{Model-specific stylistic profiles}

Although both model--workflow configurations exhibit compressed formal ranges, there is no stylistic convergence of GPT and Qwen. The lexical means of GPT are larger, the entropy of Qwen is smaller, Qwen is more readable, and the punctuation in both models is quite distinctive. Despite these differences in average values, each model--condition combination forms a denser cluster than its corresponding human corpus, particularly in sentence structure and readability.

The mean values indicate that there is no single justification for the claims about the lexical nature of AI fiction being either lexically poor or lexically rich. For example, GPT Victorian has greater MATTR and entropy compared to the historical comparison set, whereas Qwen Victorian has a similar mean MATTR value to the human one but lower entropy. GPT Zero-Style has slightly higher mean values for both lexical measures than the human Zero-Style corpus, whereas Qwen Zero-Style has lower mean values for both.

The results were also consistent with the human-authored structure. Removal of Ha Jin and Xiaolu Guo, authors who were responsible for fourteen out of the sixty-five Zero-Style novels, produced only small difference in the dispersion ratios. All eight 95\% percentile intervals from the author-equalized resampling of the two generated Zero-Style conditions stayed under 1. Within-author calibration revealed eighteen historical authors and six Zero-Style authors that were eligible for comparison; the generated sentence-length dispersion ratio was less than the observed within-author dispersion in all comparisons with median ratios between 0.138 and 0.206.

\begin{table*}[t]
\centering
\small
\caption{Mean punctuation frequencies and inter-novel dispersion ratios.}
\label{tab:5}
\renewcommand{\arraystretch}{1.08}
\resizebox{\textwidth}{!}{%
\begin{tabular}{@{}llllll@{}}
\toprule
\textbf{Comparison} & \textbf{Mark} & \textbf{Generated / 1,000 chars} & \textbf{Human / 1,000 chars} & \textbf{Mean ratio} & \textbf{SD ratio} \\
GPT Victorian / 19C Human & Semicolon & 0.763 & 1.958 & 0.389 & 0.200 \\
GPT Victorian / 19C Human & Em dash & 0.139 & 0.593 & 0.235 & 0.062 \\
GPT Victorian / 19C Human & Colon & 0.383 & 0.375 & 1.022 & 0.240 \\
GPT Victorian / 19C Human & Comma & 10.822 & 14.866 & 0.728 & 0.191 \\
Qwen Victorian / 19C Human & Semicolon & 0.020 & 1.958 & 0.010 & 0.012 \\
Qwen Victorian / 19C Human & Em dash & 1.006 & 0.593 & 1.696 & 0.064 \\
Qwen Victorian / 19C Human & Colon & 0.092 & 0.375 & 0.244 & 0.071 \\
Qwen Victorian / 19C Human & Comma & 14.933 & 14.866 & 1.005 & 0.166 \\
GPT Zero-Style / Quinn Human & Semicolon & 0.045 & 0.266 & 0.168 & 0.108 \\
GPT Zero-Style / Quinn Human & Em dash & 0.033 & 0.440 & 0.075 & 0.048 \\
GPT Zero-Style / Quinn Human & Colon & 0.537 & 0.507 & 1.060 & 0.342 \\
GPT Zero-Style / Quinn Human & Comma & 9.171 & 11.617 & 0.789 & 0.229 \\
Qwen Zero-Style / Quinn Human & Semicolon & 0.008 & 0.266 & 0.030 & 0.019 \\
Qwen Zero-Style / Quinn Human & Em dash & 0.962 & 0.440 & 2.187 & 0.213 \\
Qwen Zero-Style / Quinn Human & Colon & 0.127 & 0.507 & 0.251 & 0.087 \\
Qwen Zero-Style / Quinn Human & Comma & 14.143 & 11.617 & 1.217 & 0.361 \\
\bottomrule
\end{tabular}%
}
\end{table*}

The punctuation comparisons are descriptive, however, punctuation usage illustrates how no single stylistic signature exists for AI. GPT uses fewer semicolons and em dashes compared to both human corpora. While Qwen uses even fewer semicolons, it uses more em dashes compared to the human corpora: 1.006 per 1,000 characters for Qwen Victorian and 0.962 for Qwen Zero-Style. The Qwen Victorian comma usage approximates the historical average, whereas the Qwen Zero-Style usage surpasses the modern human average. Despite this difference in average levels, punctuation variance is substantially lower in all four novel corpora. The SD ratios range from 0.012 to 0.361.

\subsection{Exploratory cross-feature coordination}

Cross-feature coordination was examined as an exploratory extension of the variance analysis.

\begin{table*}[t]
\centering
\small
\caption{Cross-family correlations and author-balanced Fisher-z differences.}
\label{tab:6}
\renewcommand{\arraystretch}{1.08}
\resizebox{\textwidth}{!}{%
\begin{tabular}{@{}lllllll@{}}
\toprule
\textbf{Comparison} & \textbf{Pair} & \textbf{Generated Pearson r} & \textbf{Generated Spearman rho} & \textbf{Human Pearson r} & \textbf{Median z difference} & \textbf{95\% percentile interval} \\
GPT Victorian / 19C Human & MATTR-500--sentence length & 0.363 & 0.397 & 0.153 & 0.235 & -0.363--0.869 \\
GPT Victorian / 19C Human & Entropy--sentence length & 0.304 & 0.414 & 0.013 & 0.341 & -0.295--0.990 \\
Qwen Victorian / 19C Human & MATTR-500--sentence length & -0.081 & -0.110 & 0.153 & -0.247 & -0.825--0.349 \\
Qwen Victorian / 19C Human & Entropy--sentence length & -0.246 & -0.218 & 0.013 & -0.247 & -0.913--0.422 \\
GPT Zero-Style / Quinn Human & MATTR-500--sentence length & 0.253 & 0.334 & -0.058 & 0.483 & -0.121--1.101 \\
GPT Zero-Style / Quinn Human & Entropy--sentence length & 0.154 & 0.263 & -0.152 & 0.389 & -0.282--1.091 \\
Qwen Zero-Style / Quinn Human & MATTR-500--sentence length & -0.509 & -0.606 & -0.058 & -0.396 & -1.047--0.171 \\
Qwen Zero-Style / Quinn Human & Entropy--sentence length & -0.591 & -0.668 & -0.152 & -0.479 & -1.082--0.038 \\
\bottomrule
\end{tabular}%
}
\end{table*}

\emph{Note: Pearson differences are Fisher-z transformed; 95\% percentile intervals are based on 10,000 author-balanced resamples.}

The coordination of features is less consistent compared to variance compression. GPT Victorian exhibits positive correlations between sentence length and MATTR (r = .363) and entropy (r = .304); however, both Fisher-z intervals include zero. Qwen Victorian lacks any similar positive correlation; its correlations are weakly negative, and the corresponding intervals also include zero.

Zero-Style conditions are even further apart. GPT Zero-Style displays weak positive correlations between lexical and sentence measures, but both Fisher-\emph{z} intervals include zero, providing no clear evidence that these relationships differ from those in the human Zero-Style corpus. In contrast, Qwen Zero-Style displays negative correlations for MATTR (r = -.509) and entropy (r = -.591), and both corresponding Fisher-\emph{z} intervals include zero. Because the direction and strength of these relationships differ across models and stylistic conditions, the present results do not support a general pattern of correlational overclosure.

\section{Discussion}

\subsection{Variance overclosure as the principal finding}

The main finding is that variance overclosure is observed across both model--workflow configurations and both stylistic targets. GPT Victorian, Qwen Victorian, GPT Zero-Style, and Qwen Zero-Style are all located within a smaller inter-novel range than their human counterparts in terms of average sentence length, intra-novel sentence length variance, readability, and punctuation.

Cross-workflow results for sentence structure are the most compelling evidence. There are differences in the mean sentence lengths between the models, but the document-level means and intra-novel sentence lengths have very tight distributions across all four generation conditions. This finding is true even when taking into account the number of documents, adjusting for document lengths, and restricting the historical comparison to novels published during the Victorian period.

The conditions for novel generation are not perfectly matched. GPT Zero-Style prompts focus on migration, diaspora, intercultural organizations, or minorities, while Qwen uses two style locks for each neutral story seed. The GPT Victorian and Qwen Victorian protocols also vary in text length, prompt delivery, structure of chapters, and context handling. This difference broadens the cross-workflow comparison but prevents a clear factorial analysis.

Sentence-length compression arises because sentence rhythm reflects recurring local generation preferences reinforced by shared prompts, continuation procedures, and context-management conventions. Lexical diversity, by contrast, accumulates across changing topics, settings, and characters. This interpretation is limited by the design' s inability to separate model architecture from prompting, decoding, segmentation, or context management.

Overclosure should consequently be viewed as related yet different from other types of AI homogenization. Diversity in literature does not only appear through mean attributes but also through unusual combinations and formal outliers. A generation process that consistently cycles back to limited ranges can maintain fluency yet limit formal options. The recurring pattern is therefore not convergence on identical prose, but repeated occupation of unusually narrow formal ranges.

Human-author sensitivity analyses, however, make clear that sentence-length compression is not only due to comparing one generation workflow to the multi-author corpus: All four conditions showed lower sentence-length dispersion than every eligible human-author sample examined. The less consistent lexical and readability results identify sentence length as the most robust dimension of variance overclosure.

\subsection{Different models, different stylistic locations}

Variance compression does not imply that GPT and Qwen produce the same style. Their mean lexical values, readability levels, sentence lengths, and punctuation profiles differ substantially. Lexical evidence sharpens this claim. Novels generated by the two models are not lexically inferior across the board, and there is no lexical average level shared by the two models. GPT surpasses the human means in both conditions; Qwen Victorian is close to the human MATTR mean but has lower entropy; and Qwen Zero-Style is low on both.

Overclosure is not an indicator of literary merit, and this analysis does not evaluate originality, impact, complexity, reader reaction, or the aesthetics of literature. The method cannot be utilized as a stand-alone means for the detection of any disputed works either, but is intended for known collections and comparative studies only.

Neither should overclosure be defined by low values of each metric. The GPT Victorian model utilizes a higher level of lexical sophistication, whereas the Qwen Victorian model stays close to the human mean for MATTR and is compressed. The key evidence for variance overclosure is distributional: the amount of variation among novels within a corpus.

\subsection{Correlational overclosure as a conditional hypothesis}

Correlational overclosure remains a conditional and exploratory concept. The six-corpus approach shows that differences between GPT and Qwen are observed in both the values and the strengths of the correlations, with the most notable unusual correlations manifesting themselves in the form of negative lexical--sentence correlations in Qwen Zero-Style. Broader claims will require more model families, matched and independent prompt sets, prespecified correlation families, and larger numbers of independently generated documents.

\subsection{Qualitative counterpoint: what the metrics do not capture}

The statistical results describe distributions across novels, but they do not reveal the properties of the isolated passages which are immediately perceptible during the act of reading. In other words, is overclosure perceptible when we read these AI-generated novels? In general, no. We read three AI-generated novels to see whether differences in metrics are observable. The novels were \emph{Where the Bell Calls}, \emph{The House Between Two Tongues}, and \emph{The Quota Line} -- the first two from the Victorian group, the third from the Zero-Style group.

Within the corpus of AI Victorian novels, \emph{Where the Bell Calls} and \emph{The House Between Two Tongues} lie at either end of the scale for lexical diversity. For lexical density, they are close; the Flesch values for this corpus range from 62 to 72, placing the two novels on either side of the average for readability. We chose \emph{The Quota Line} from the Zero-Style AI corpus because it is set in contemporary Ireland, where one of us was born and raised. We were curious whether a reader's knowledge of the setting might enhance the analysis.

The differences in lexical diversity and lexical density between a real nineteenth-century novel and an AI-authored one seem subliminal, or at least below our thresholds as readers. The chapter openings of the AI-generated nineteenth-century novels were particularly impressive, displaying a mastery of lexis and syntax that equalled anything in the canon. Here is the opening paragraph of \emph{The House Between Two Tongues}, set on the border of England and Wales:

\begin{quote}
Miss Clara Ashcombe first saw Durnford March from the high road above the Celyn valley, where the coach, having laboured for some miles through a country of wet hedges, stony pastures, and leafless orchards, paused while the horses were watered at a wayside trough. It was one of those February afternoons in which the light seems unwilling either to declare itself day or resign itself to evening. The sky hung low and pale over the fields; the distant hills, which belonged already more to Wales than to England, were blurred by a faint vapour; and below, in the hollow where the river bent round a shoulder of meadowland, the town lay gathered about its bridge, its church-tower, its market roofs, and the darker line of houses that climbed toward the chapel quarter.
\end{quote}

The scene anchors the focalizer, Clara Ashcombe, and also deftly foreshadows the novel's bilingual theme, as a ``faint vapour'' blurs the boundary between England and Wales, which the novel's plot will do in more detail. It is evident that the overclosure of lexical diversity, density, readability, sentence length, and punctuation rhythm, revealed by statistical analysis, is difficult, if not impossible, for a reader to perceive. When encountering such paragraphs, we often felt that we were reading recently unearthed classics of Victorian realism.

However, a different aspect of the novels \emph{was} perceptible, but whether this reflects the deep structure we describe is debatable. One of the most striking features of the three AI-generated novels examined here is their wit. Across the three novels, we find dialogues that sparkle with bracing, mordant concision. It is not drollery, and it is not satire; rather, characters sum up particular relationships or social contexts with declarations that are both direct and imaginatively figurative. Here is an example from the GPT Victorian corpus:

\begin{quote}
Rose, meanwhile, had been drawn into conversation by Sybil Mowbray, who asked whether she still painted.

``When I have paper,'' said Rose.

Sybil, uncertain whether this was pathos or impertinence, answered kindly, ``That must be a great pleasure.''

``Paper?''

``Painting.''

``It is sometimes a pleasure. Sometimes it is only a way of putting colours where life has left none.'' (\emph{Where the Bell Calls})
\end{quote}

Unsure whether Rose speaks with feeling or rudeness, Sybil gives her the benefit of the doubt, which is kind but condescending. There is stylistic delicacy in the depiction of Sybil's disposition toward Rose, which is structural and not personal, i.e., it results not from Sybil's particular feeling toward Rose but from their different social positions. Thus, the hesitation also tellingly reveals the characters' difference in social status. The exchange is all the more pleasurable when, after Sybil graciously acknowledges Rose's pleasure, Rose is victorious, as she captures the constrictions on her life through the figure of painting introduced by her interlocutor. It is comparable to the exchanges in novels like \emph{Pride and Prejudice} and \emph{Jane Eyre}.

Two examples of dialogue best capture another related aspect of these novels. Here, one of the speakers extrapolates the details of a proximate situation to make a far-reaching generalisation. The first is from the Victorian corpus:

\begin{quote}
``This list is not finished,'' she said. ``That is its virtue.''

Edith looked startled, then pleased. ``Most school exercises are meant to be finished.''

``Most school exercises pretend knowledge is tidier than it is.''

Rosamund looked at her fair copy with some alarm. ``Will it become messy?''

``Yes,'' Miss Rees said. ``If it becomes useful.'' (\emph{The House Between Two Tongues})
\end{quote}

Education is a recurring theme in the three AI novels we read, and here Miss Rees's equation of the messy and the useful offers insight into the process of learning. It is witty because it is counterintuitive: surely schools shouldn't teach children to be messy. The next is from the Zero-Style AI corpus:

\begin{quote}
``You'll forget to eat,'' she said.

``I had tea.''

``Tea is not food.''

``It has kept this country alive.''

``It has kept this country irritable.'' (\emph{The Quota Line})
\end{quote}

With the statement about keeping the country irritable, Catriona deftly dissects her husband's remark to make a grand declaration about Ireland, which may ultimately refer to her husband's present state of mind. It is funny---a sweeping generalisation that also applies to the proximate scene.

We dwell at length on these exchanges not because they are outstanding, but because they recur across the three novels examined closely. (A cursory glance at other novels in the two AI corpora confirms this.) The ubiquity of such wit results in stylistic monotony, as there are few other modes of dialogue. (It seemed particularly out of place in \emph{The Quota Line}, which is set in an Irish village in the mid to late 2000s. Such communities have a varied palette of wit, but this is not one of them.)

We also frequently encountered zeugma in the three novels examined closely, or, more properly, one type of zeugma, syllepsis. Katy Waldman has remarked on its use, along with negative parallelism, in AI fiction. While we found little of the latter in the three novels, zeugma is prevalent in the novels' dialogues. Waldman continues:

\begin{quote}
These rhetorical devices exploit our learned associations between certain types of repetition in prose and heightened meaning; they also create a pulse we feel in our bodies. If they recur in automated text, it's because they recur in human writing, but in the fake stuff they're decoupled from content. (Waldman 2026)
\end{quote}

Is Rose's wit above ``decoupled from content''? No, it emerges organically from the social situation, as our analysis showed. But when so many characters display rhetorical skills comparable to Rose, we might feel that this is an inaccurate representation of conversations. As in pornography, where all exchanges, however trivial, lead to sex, characters in these novels cannot even chat about the weather without Wildean sprezzatura. While this effect is immediately manifest to a reader, it is unlikely to affect the metrics we have applied to our six corpora.

Marriage was perhaps the most important plot structure in nineteenth-century Anglophone fiction. It is noteworthy, then, that in the Victorian AI novel, \emph{Where the Bell Calls}, the protagonists get married about a third of the way through the book. Another popular plot device in Victorian fiction (and, obviously, in much other literature) is the paragon who turns out to be an evil antagonist (for instance, \emph{Dr Jekyll and Mr Hyde} or \emph{The Picture of Dorian Gray}). In \emph{The House Between Two Tongues}, a charming villain is set up at the outset (Mr Trevask), only to be unmasked about a third of the way through (around the same point as the wedding in \emph{Where the Bell Calls}); he is then parked offstage for the remainder of the book, denying readers the pleasure of a denouement or showdown. In these two AI novels, the A plots no longer drive the stories, and we might say that halfway through, the protagonists drift into episodic B plots. This is not necessarily a stylistic failing; arguably, \emph{Middlemarch} excels because it, too, solves the marriage issue early on and then examines its effects. However, these two novels do not sustain the interlaced plots associated with nineteenth-century fiction over 80,000 to 120,000 words and settles into an episodic organisation that can easily end at any point. Once again, it is unlikely that this structural feature affects the metrics we have applied here.

Perhaps related to this, all three novels reached solutions that involved community negotiations and the establishment of organic institutional structures that might be labelled cooperatives. This was notable because none of the prompts included such a thematic instruction.

\section{Limitations}

First of all, there is a common set of factors associated with the generation of each group of novels, namely, the prompts used, the settings adopted, the workflow followed, and the period of generation; furthermore, there are other characteristics like model identity, interface, decoding, segmentation, chapter goal, and context management that are consistently varying between the two workflows.

Second, the human comparator group contains multiple authors, whereas each AI corpus has repeated generations by a single workflow. Calibration within an author partially resolves this imbalance; however, only eighteen historical and six Zero-Style authors produced three or more novels. Canon and digitization bias also exist in the historical corpus, and the human Zero-Style corpus is not evenly represented among its authors.

Finally, the measures capture only selected surface properties rather than literary form in its entirety. They do not assess literary features such as plot, characterization, or literary value.

\section{Conclusion}

This paper proposes the term ``overclosure'' to describe a pattern seen at the corpus level in AI-generated long fiction. Across the two model--workflow configurations and stylistic targets examined, the generated novels showed narrower document-level formal ranges than the human corpora. The clearest evidence can be seen in sentence-length dispersion which remains unusually low and compressed even after author-balanced resampling, length adjustment, removing each author individually to check their effect, and restricting the comparison group to just the Victorian period. Readability, punctuation, and intra-novel sentence-length variation provide further evidence, while the lexical results are more conditional, particularly for Qwen Zero-Style MATTR. The result is limited to the observed workflows and selected human comparators and should not be interpreted as a universal estimate of language-model fiction.

The compressions do not produce a single AI writing style. GPT and Qwen display very diverse values for mean lexical diversity, entropy, sentence length, reading levels, and punctuation distribution. Therefore, they do not share the same style of writing and each has its own average values. Consequently, overclosure should be understood as reduced formal range, rather than having low metric values or similar style characteristics.

For cross-feature correlations, the evidence is less consistent. There are correlations between lexical variables and sentence length, but they vary from one model to another as well as among different stylistic scenarios. Moreover, there are no common correlation patterns that could be observed across all four AI corpora. Therefore, there is no support for a general approach to the phenomenon of correlational overclosure.

\section*{Data availability}

The replication package includes the Victorian and Zero-Style prompt files, four generated corpus files, Qwen generation, configuration information, Qwen corpus files, document-level feature tables, and statistics. Supplementary Data 1 consists of one row per novel with corpus, model or author, bibliographic identification if available, word count, and all other metrics calculated. The document-level data, generated corpora, prompts, and replication code supporting this study are available in the OSF repository at \href{https://osf.io/8akph/overview?view_only=074589cee59740c4896b9c409cc54ea9}{this link}. Copyrighted contemporary human novels are not redistributed; reproducing feature extraction for those texts requires lawful access to the source novels.''

\section*{Author Contributions}

M.S.P. conceived the study, generated the model corpora, conducted the computational analyses, and drafted the quantitative sections. J.Q. selected the Zero-Style human corpus, conducted the close reading, contributed to the interpretation, and revised the manuscript. Both authors reviewed the results and approved the final manuscript.

\section*{Ethics statement}

The study analyzes published literary texts and model-generated fiction. It does not involve human participants, private personal data, or intervention with human subjects.

\section*{Competing interests}

The authors declare no competing interests.

\section*{AI-use disclosure}

The ChatGPT interface and GPT-5.5 Thinking were used to generate the two AI fiction corpora. In the supplementary material section, you can see the prompt transcription and original raw prompt files. ChatGPT was also used for drafting and debugging analysis code and language editing. The authors selected and curated the corpora, designed the study, reviewed and executed the analyses, checked the code and outputs, verified citations and reported values. The authors remain fully responsible for the interpretation and final manuscript.


\begin{thebibliography}{99}

\bibitem[Agarwal et al.(2025)]{ref01} Agarwal D, Naaman M, Vashistha A (2025) AI suggestions homogenize writing toward Western styles and diminish cultural nuances. In: Proceedings of the 2025 CHI Conference on Human Factors in Computing Systems. Association for Computing Machinery, New York, Article 1117, pp 1--21. \url{https://doi.org/10.1145/3706598.3713564}

\bibitem[Anderson et al.(2024)]{ref02} Anderson BR, Shah JH, Kreminski M (2024) Homogenization effects of large language models on human creative ideation. In: Proceedings of the 16th Conference on Creativity \& Cognition. Association for Computing Machinery, New York, pp 413--425. \url{https://doi.org/10.1145/3635636.3656204}

\bibitem[Bajohr(2024)]{ref03} Bajohr H (2024) Cohesion without coherence: artificial intelligence and narrative form. TRANSIT 14(2). \url{https://doi.org/10.5070/T714264658}

\bibitem[Benjamini et al.(1995)]{ref04} Benjamini Y, Hochberg Y (1995) Controlling the false discovery rate: a practical and powerful approach to multiple testing. Journal of the Royal Statistical Society: Series B 57:289--300

\bibitem[Biber et al.(2009)]{ref05} Biber D, Conrad S (2009) Register, genre, and style. Cambridge University Press, Cambridge

\bibitem[Brown et al.(1974)]{ref06} Brown MB, Forsythe AB (1974) Robust tests for the equality of variances. Journal of the American Statistical Association 69:364--367

\bibitem[Ciphers of the Times(2022)]{ref07} Ciphers of the Times (2022) Victorian novels corpus. McGill University Library. \url{https://libraryponders.github.io/resources.html}. Accessed 4 Aug 2026

\bibitem[Coleman et al.(1975)]{ref08} Coleman M, Liau TL (1975) A computer readability formula designed for machine scoring. Journal of Applied Psychology 60:283--284

\bibitem[Covington et al.(2010)]{ref09} Covington MA, McFall JD (2010) Cutting the Gordian knot: the moving-average type--token ratio (MATTR). Journal of Quantitative Linguistics 17:94--100

\bibitem[Crossley et al.(2008)]{ref10} Crossley SA, Greenfield J, McNamara DS (2008) Assessing text readability using cognitively based indices. TESOL Quarterly 42:475--493

\bibitem[Dale et al.(1948)]{ref11} Dale E, Chall J (1948) A formula for predicting readability. Educational Research Bulletin 27:11--28

\bibitem[Darmon et al.(2021)]{ref12} Darmon ANM, Bazzi M, Howison SD, Porter MA (2021) Pull out all the stops: textual analysis via punctuation sequences. European Journal of Applied Mathematics 32:1069--1105. \url{https://doi.org/10.1017/S0956792520000157}

\bibitem[Eder(2017)]{ref13} Eder M (2017) Short samples in authorship attribution: a new approach. Digital Scholarship in the Humanities 32:294--305

\bibitem[Eder et al.(2016)]{ref14} Eder M, Rybicki J, Hoover DL (2016) Computational stylistics and text analysis. In: Crompton C, Lane RJ, Siemens R (eds) Doing digital humanities. Routledge, London, pp 123--144

\bibitem[Efron et al.(1993)]{ref15} Efron B, Tibshirani RJ (1993) An introduction to the bootstrap. Chapman \& Hall, New York

\bibitem[Flesch(1948)]{ref16} Flesch R (1948) A new readability yardstick. Journal of Applied Psychology 32:221--233

\bibitem[Gass(1985)]{ref17} Gass WH (1985) Habitations of the word: essays. Simon \& Schuster, New York

\bibitem[Gunning(1952)]{ref18} Gunning R (1952) The technique of clear writing. McGraw-Hill, New York

\bibitem[Jockers(2013)]{ref19} Jockers ML (2013) Macroanalysis: digital methods and literary history. University of Illinois Press, Urbana

\bibitem[Kincaid et al.(1975)]{ref20} Kincaid JP, Fishburne RP Jr, Rogers RL, Chissom BS (1975) Derivation of new readability formulas for Navy enlisted personnel. Naval Technical Training Command, Millington

\bibitem[Kobak et al.(2025)]{ref21} Kobak D, González-Márquez R, Horvát E-Á, Lause J (2025) Delving into LLM-assisted writing in biomedical publications through excess vocabulary. Science Advances 11:eadt3813. \url{https://doi.org/10.1126/sciadv.adt3813}

\bibitem[McLaughlin(1969)]{ref22} McLaughlin GH (1969) SMOG grading---a new readability formula. Journal of Reading 12:639--646

\bibitem[Moretti(2013)]{ref23} Moretti F (2013) Distant reading. Verso, London

\bibitem[Nunberg(1990)]{ref24} Nunberg G (1990) The linguistics of punctuation. CSLI Lecture Notes, vol 18. Center for the Study of Language and Information, Stanford

\bibitem[O'Sullivan(2025)]{ref25} O'Sullivan J (2025) Stylometric comparisons of human versus AI-generated creative writing. Humanities and Social Sciences Communications 12:1708. https://doi.org/10.1057/s41599-025-05986-3

\bibitem[Pawłowski et al.(2026)]{ref26} Pawłowski A, Walkowiak T (2026) Stylometric approach to AI-generated texts: an analysis of contemporary French-language literature. In: Proceedings of the 10th Joint SIGHUM Workshop on Computational Linguistics for Cultural Heritage, Social Sciences, Humanities and Literature. Association for Computational Linguistics, Rabat, pp 221--226. \url{https://aclanthology.org/2026.latechclfl-1.21/}

\bibitem[Pitler et al.(2008)]{ref27} Pitler E, Nenkova A (2008) Revisiting readability: a unified framework for predicting text quality. In: Proceedings of the 2008 Conference on Empirical Methods in Natural Language Processing. Association for Computational Linguistics, Honolulu, pp 186--195

\bibitem[Przystalski et al.(2026)]{ref28} Przystalski K, Argasiński JK, Grabska-Gradzińska I, Ochab JK (2026) Stylometry recognizes human and LLM-generated texts in short samples. Expert Systems with Applications 296:129001. https://doi.org/10.1016/j.eswa.2025.129001

\bibitem[Quinn(2025)]{ref29} Quinn J (2025) Literature in the age of lingua franca English: the zero style. Routledge, London. https://doi.org/10.4324/9781003521587

\bibitem[Rettberg et al.(2025)]{ref30} Rettberg JW, Wigers H (2025) AI-generated stories favour stability over change: homogeneity and cultural stereotyping in narratives generated by GPT-4o-mini. Open Research Europe 5:202. https://doi.org/10.12688/openreseurope.20576.1

\bibitem[Senter et al.(1967)]{ref31} Senter RJ, Smith EA (1967) Automated readability index. Aerospace Medical Research Laboratories, Wright-Patterson Air Force Base

\bibitem[Shannon(1948)]{ref32} Shannon CE (1948) A mathematical theory of communication. Bell System Technical Journal 27:379--423

\bibitem[Stamatatos(2009)]{ref33} Stamatatos E (2009) A survey of modern authorship attribution methods. Journal of the American Society for Information Science and Technology 60:538--556

\bibitem[Tanaka-Ishii et al.(2015)]{ref34} Tanaka-Ishii K, Aihara S (2015) Computational constancy measures of texts---Yule's K and Rényi's entropy. Computational Linguistics 41:481--502

\bibitem[Underwood(2019)]{ref35} Underwood T (2019) Distant horizons: digital evidence and literary change. University of Chicago Press, Chicago

\bibitem[Vachek(1989)]{ref36} Vachek J (1989) Written language revisited. Luelsdorff PA (ed). John Benjamins, Amsterdam. \url{https://doi.org/10.1075/z.41}

\bibitem[Waldman(2026)]{ref37} Waldman, Katy. ``Did a Chatbot Write a Prize-Winning Story? Does It Matter?'' \emph{The New Yorker}, 10 June 2026, \href{http://www.newyorker.com/books/page-turner/did-a-chatbot-write-a-prize-winning-story-does-it-matter}{New Yorker online}

\bibitem[Wilkens(2026)]{ref38} Wilkens M (2026) AI as a tool for simulation-based experiments in literary studies. arXiv preprint arXiv:2606.02293. https://doi.org/10.48550/arXiv.2606.02293

\bibitem[Yang et al.(2025)]{ref39} Yang A, Li A, Yang B et al (2025) Qwen3 technical report. arXiv preprint arXiv:2505.09388. https://doi.org/10.48550/arXiv.2505.09388

\end{thebibliography}
\end{document}